\def\paperID{0000}
\def\confName{CVPR}
\def\confYear{2024}

\def\paperTitle{Iterative Refinement Strategy for Automated Data Labeling: Facial Landmark Diagnosis in Medical Imaging}

\def\authorBlock{
    Yu-Hsi Chen \\
    The University of Melbourne \\
    {\tt\small yuhsi@student.unimelb.edu.au}
}

\newif\ifreview 
\newif\ifarxiv 
\newif\ifcamera \newcommand{\cameraready}{\cameratrue}
\newif\ifrebuttal 
\cameraready 

\pdfoutput=1
\documentclass[10pt,twocolumn,letterpaper]{article}
\ifreview \usepackage[review]{cvpr} \fi
\ifarxiv \usepackage[pagenumbers]{cvpr} \fi
\ifrebuttal \usepackage[rebuttal]{cvpr} \fi
\ifcamera \usepackage{cvpr} \fi

\usepackage{graphicx}	
\usepackage{amsmath}	
\usepackage{amssymb}	
\usepackage{booktabs}
\usepackage{times}
\usepackage{microtype}
\usepackage{epsfig}
\usepackage[table,xcdraw,dvipsnames]{xcolor}
\usepackage{caption}
\usepackage{float}
\usepackage{placeins}
\usepackage{color, colortbl}
\usepackage{stfloats}
\usepackage{enumitem}
\usepackage{tabularx}
\usepackage{xstring}
\usepackage{multirow}
\usepackage{xspace}
\usepackage{url}
\usepackage{subcaption}
\usepackage{xcolor}
\usepackage{algorithm}
\usepackage{algorithmic}  
\usepackage[hang,flushmargin]{footmisc}

\ifcamera \usepackage[accsupp]{axessibility} \fi

% ...

\ifarxiv  \fi

\newcommand{\R}[1]{{%
    \textbf{%
        \ifstrequal{#1}{1}{\textcolor{red}{R#1}}{%
        \ifstrequal{#1}{2}{\textcolor{blue}{R#1}}{%
        \ifstrequal{#1}{3}{\textcolor{magenta}{R#1}}{%
        \ifstrequal{#1}{4}{\textcolor{teal}{R#1}}{%
                           \textcolor{cyan}{R#1}%
        }}}}%
    }%
}}

\usepackage{xr-hyper}

\makeatletter
\newcommand*{\addFileDependency}[1]{
  \typeout{(#1)}
  \@addtofilelist{#1}
  \IfFileExists{#1}{}{\typeout{No file #1.}}
}

\makeatother

\definecolor{cvprblue}{rgb}{0.21,0.49,0.74}
\usepackage[pagebackref,breaklinks,colorlinks,citecolor=cvprblue]{hyperref}
\usepackage[capitalize]{cleveref}
\crefname{section}{Sec.}{Secs.}
\crefname{table}{Table}{Tables}
\crefname{figure}{Fig.}{Figs.}

\begin{document}
\title{\paperTitle}
\author{\authorBlock}
\maketitle

\begin{abstract}
Automated data labeling techniques are crucial for accelerating the development of deep learning models, particularly in complex medical imaging applications. However, ensuring accuracy and efficiency remains challenging. This paper presents iterative refinement strategies for automated data labeling in facial landmark diagnosis to enhance accuracy and efficiency for deep learning models in medical applications, including dermatology, plastic surgery, and ophthalmology. Leveraging feedback mechanisms and advanced algorithms, our approach iteratively refines initial labels, reducing reliance on manual intervention while improving label quality. Through empirical evaluation and case studies, we demonstrate the effectiveness of our proposed strategies in deep learning tasks across medical imaging domains. Our results highlight the importance of iterative refinement in automated data labeling to enhance the capabilities of deep learning systems in medical imaging applications. 
\cameraready
The source code is available in \url{https://github.com/wish44165/iAutolabeling}.

\end{abstract}
\section{Introduction}
\label{sec:intro}

\quad Acquiring accurately labeled datasets is essential for developing robust and applicable models in deep learning applications. However, conventional manual annotation methods present inherent limitations, such as labor intensiveness, time consumption, and susceptibility to human errors. These challenges have prompted the exploration of alternative approaches, leading to the emergence of automated data labeling techniques. These have garnered increasing interest within the research community due to their potential to streamline the labeling process, mitigate the associated labor burdens, and enhance overall efficiency. By utilizing sophisticated algorithms and machine learning frameworks, automated data labeling methods strive to enhance annotation accuracy while minimizing manual labor, thereby facilitating the generation of high-quality datasets crucial for practical model training and deployment across various domains.

\begin{figure}[tp]
    \centering
    \includegraphics[width=\linewidth]{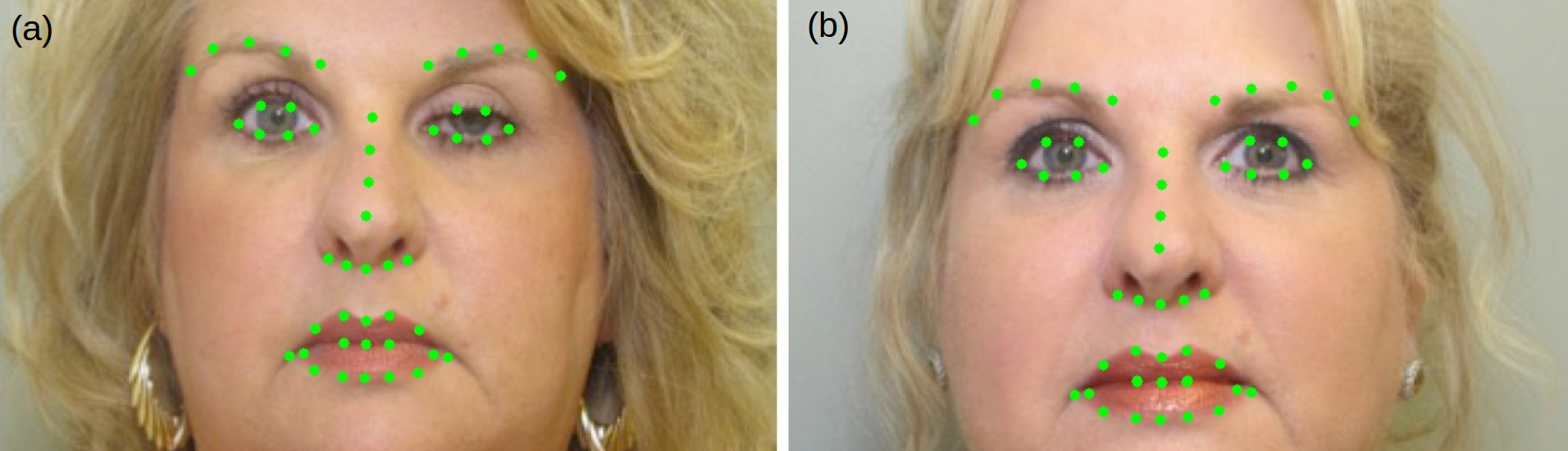}
    \caption{The appearance of eyelid problems is provided in~\cite{farber2020evaluation}, and the green dots indicate the 51 keypoints used for annotating facial landmarks. (a) Patient with acquired left ptosis after left upper lid hematoma treated conservatively for a year. (b) The patient had a left tarsoaponeurectomy and was slightly overcorrected, which can easily be treated with downward lid massage.}
    \label{fig:ptosis_example}
\end{figure}

Despite the promise of automated labeling methods, concerns persist regarding the accuracy and reliability of generated labels, particularly in the context of complex deep learning tasks where precise annotations are indispensable for optimal model performance. Addressing this critical concern necessitates developing and exploring innovative strategies that enhance the accuracy and efficiency of automated data labeling processes.
Moreover, the significance of facial keypoint detection extends to various medical imaging domains, including dermatology, plastic surgery, and ophthalmology. In dermatology, automated facial keypoint detection assists in analyzing skin conditions, lesions, and abnormalities by identifying key facial landmarks such as moles, wrinkles, and discolorations. Similarly, precise facial keypoint detection in plastic surgery facilitates preoperative planning, surgical simulation, and postoperative evaluation, enabling surgeons to achieve desired aesthetic outcomes and minimize complications. In ophthalmology~\cite{hung2022outperforming,farber2020evaluation}, facial keypoint detection aids in diagnosing and monitoring eye conditions such as strabismus and ptosis, as shown in Figure~\ref{fig:ptosis_example}, green dots let us observe the symptom of ptosis clearly, providing insights into ocular alignment and function. Automated systems leveraging facial keypoint detection in these medical specialties offer opportunities for quantitative analysis, longitudinal monitoring, and optimization of treatment interventions.

This paper aims to bridge this gap by delving into the development and examination of iterative refinement strategies explicitly tailored for automated data labeling. The central objective revolves around refining initial labels iteratively, leveraging feedback mechanisms and sophisticated algorithms to augment the accuracy and efficiency of data labeling processes within deep learning applications. This study endeavors to ascertain the effectiveness of iterative refinement strategies by conducting a comprehensive empirical evaluation across diverse deep learning tasks, aiming to provide thorough insights into their efficacy and applicability in different scenarios. Furthermore, case studies will be conducted to elucidate these strategies' practical implications and potential benefits in real-world scenarios. By illuminating their ability to accelerate model training and enhance the overall performance of deep learning systems, this research aims to advance automated data labeling techniques in modern deep learning applications.

The contributions of this paper are outlined as follows:
\begin{itemize}
    \item[$1.$] We designed an iterative refinement strategy for an automated data labeling algorithm to provide more comprehensive and higher-quality facial keypoint label annotations.
    \item[$2.$] The feasibility of the proposed algorithm was validated through the observation of evaluation metrics during training and its high performance in real-world scenarios.
\end{itemize}
\section{Related Work}
\label{sec:related}

\subsection{Medical Imaging Analysis}
\label{ssec:medical_imaging_analysis}

\quad Medical imaging analysis, an interdisciplinary field merging computer science, engineering, and medicine, is dedicated to extracting valuable insights from medical images obtained through various modalities like X-ray, MRI, CT, and ultrasound. Its significance is aiding disease diagnosis, treatment planning, surgical guidance, and patient health monitoring. Medical imaging analysis strives to improve diagnostic precision, effectiveness, and dependability by leveraging advanced image processing techniques, machine learning algorithms, and deep learning models. Research efforts have yielded notable advancements, such as deep learning architectures for precise segmentation of anatomical structures and lesions, exemplified by studies like~\cite{huang2022fully,ronneberger2015u}. Additionally, computer-aided diagnosis systems employ machine learning and image analysis to assist radiologists, as seen in~\cite{sechopoulos2021artificial}. There is also a growing interest in integrating information from multiple imaging modalities, as discussed in~\cite{pei2023review}. Finally, translational research aims to bridge the gap between algorithm development and clinical application, with studies evaluating the real-world effectiveness of AI-driven medical imaging analysis systems. Medical imaging analysis continually evolves through these endeavors to improve patient outcomes and healthcare delivery.

\subsection{Facial Landmark Detection and Diagnosis}
\label{ssec:facial_landmark_detection_and_diagonsis}

\quad Facial landmark detection and diagnosis represents a convergence of computer vision and medical imaging domains, with a central focus on identifying critical facial points and assessing their significance across diverse applications. These landmarks are pivotal markers for understanding facial structure and identifying associated conditions, from genetic syndromes to craniofacial anomalies. By utilizing advanced image processing techniques and machine learning algorithms, this field significantly impacts clinical practice and academic research across various domains, including plastic surgery, orthodontics, dermatology, and others. The meticulous localization and analysis of facial landmarks play a pivotal role in tasks like facial expression recognition, biometric identification, and the development of personalized medical treatment strategies. Recent research has seen a surge in developing deep learning-based approaches for robust and precise facial landmark detection. For instance, Stacked Hourglass Networks for Facial Landmark Localization in~\cite{yang2017stacked}, demonstrating state-of-the-art performance in this domain.
Additionally, studies explore clinical applications of facial landmark detection in diagnosing genetic syndromes and craniofacial anomalies, as demonstrated in~\cite{cerrolaza2016identification,hennocq2023automatic}. Furthermore, facial landmark analysis is applied in cosmetic surgery planning to assess preoperative facial aesthetics and simulate surgical outcomes, as examined by~\cite{tirrell2022facial}. Automated systems integrating facial landmark detection with diagnostic algorithms are also being developed to streamline diagnosis processes, as evidenced by~\cite{kim2021automatic}. Through these research efforts, facial landmark detection and diagnosis continue to evolve, offering valuable insights into facial morphology, pathology, and treatment outcomes.

\subsection{Automated Data Labeling Techniques}
\label{ssec:automated_data_labeling_techniques}

\quad Automated data labeling techniques revolutionize annotation, offering efficient solutions for labeling large datasets. These methodologies alleviate the laborious manual effort associated with data annotation and enhance the accuracy and scalability of machine learning models. Notable research includes Activate Learning (AL) described in~\cite{ren2021survey}, which provides an extensive overview of active learning techniques in automated data labeling. AL strategies enable models to select the most informative data points for annotation, thereby optimizing the labeling process and improving model performance with minimal human intervention. Another significant work is Semi-Supervised Learning (SSL)~\cite{learning2006semi}, which delves into the category of automated labeling techniques that leverage labeled and unlabeled data to train models efficiently. By exploiting the abundance of unlabeled data, semi-supervised learning methods offer cost-effective data annotation and model training solutions, particularly when labeled data is scarce or expensive.
Additionally, Self-Supervised Learning (SSL)~\cite{jaiswal2020survey} sheds light on the effectiveness of automating data labeling tasks. SSL techniques leverage intrinsic data properties to generate supervision signals, eliminating the need for manual annotations and facilitating the training of deep learning models on large-scale datasets. Automated data labeling has played a pivotal role in advancing machine learning methodologies and paving the way for more efficient and scalable AI systems.
\section{Method}
\label{sec:method}

\quad We adhere closely to the prescribed guidelines for utilizing the official YOLOv8~\cite{Jocher_Ultralytics_YOLO_2023} pose model. A vital advantage of this simple setup is its almost plug-and-play capability.

\subsection{Automated Data Labeling}
\label{ssec:iterative_refinement_strategy}

\quad In our approach, we utilized a pre-trained model derived from the previous iteration and its predictive capabilities on the existing training dataset. This enabled us to leverage the model's predictions to generate new training instances for automated data labeling. By employing this method, the model can identify patterns and correlations within the data, facilitating the automatic labeling of additional examples. Including these newly generated training instances in the dataset enables the model to refine its understanding of the underlying data distribution and improve its predictive performance. Notably, a confidence threshold of 0.7 was employed during the prediction phase for the training dataset. This iterative process of prediction and automated labeling continuously enhances the model's accuracy and generalization ability, rendering it more effective for real-world applications.

\subsection{Non-maximum Suppression Filtering}
\label{ssec:non-maximum_suppression_filtering}

\quad Upon completing the automated labeling process, it becomes imperative to undertake a filtering mechanism to eliminate misleading or redundant labels, thereby ensuring the acquisition of comprehensive and high-quality annotation information. A threshold for non-maximum suppression (NMS)~\cite{neubeck2006efficient} is set at 0.3 to generate superiorly filtered samples. Moreover, we will prioritize the retention of original labels during the procedure. This strategic decision not only contributes to the preservation of valuable information but also bolsters the annotated dataset's overall reliability and accuracy. By retaining the original labels, we ensure that the annotated data maintains its integrity and fidelity to the ground truth. This meticulous approach ultimately fosters more robust outcomes in subsequent analyses and applications, as researchers can confidently rely on the authenticity and comprehensiveness of the dataset.

\subsection{Iterative Refinement Strategy}
\label{ssec:iterative_refinement_strategy}

\begin{figure*}[tp]
    \centering
    \includegraphics[width=\linewidth]{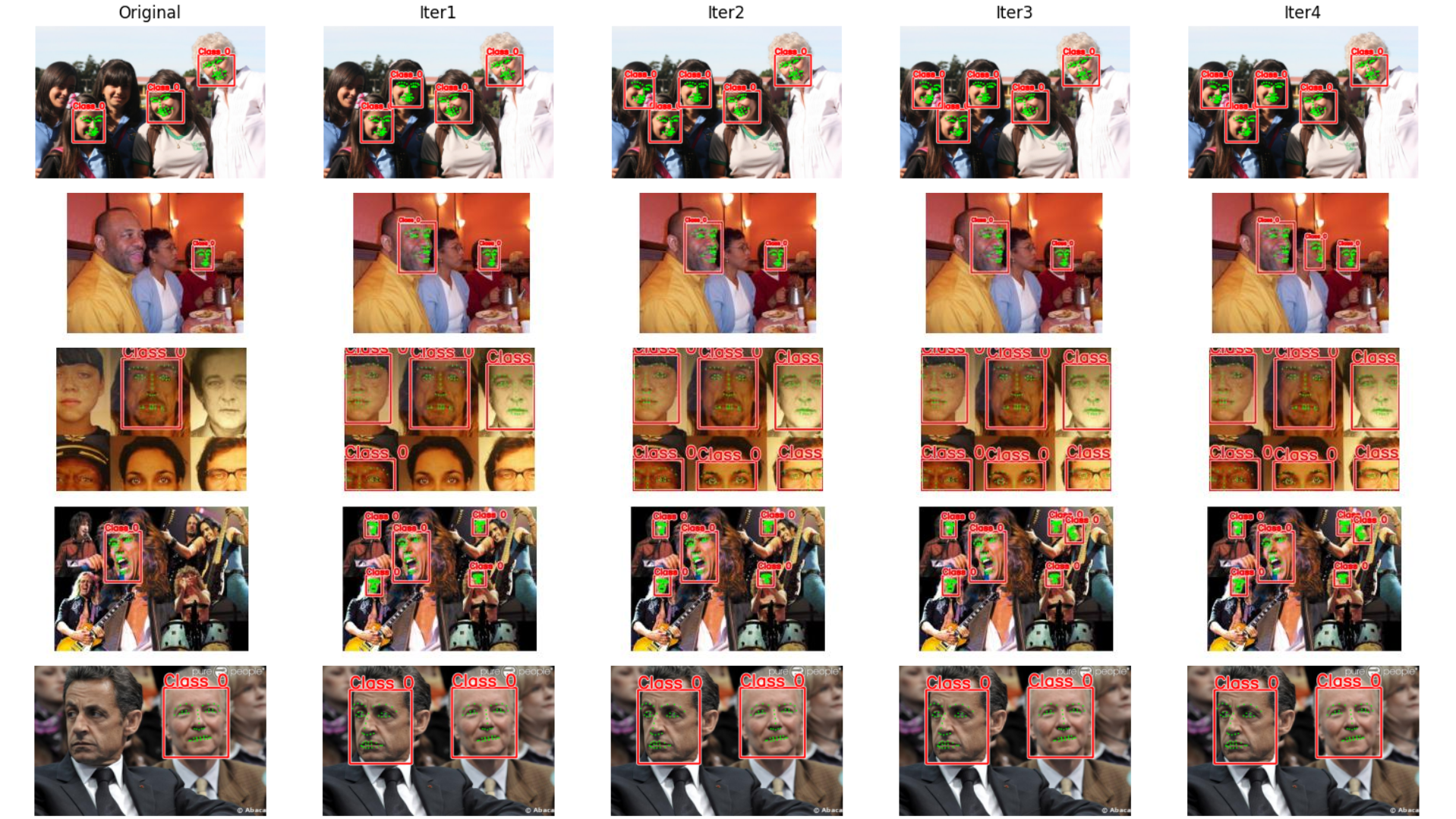}
    \caption{The horizontal axis illustrates the provided labeling information and the generated labels for each iteration, and the vertical axis denotes the images in respective datasets, namely 300W, AFW, HELEN, IBUG, and IFPW~\cite{sagonas2013300,sagonas2013semi,sagonas2016300,Zhu2012FaceDP,Le2012InteractiveFF}.}
    \label{fig:fig2}
\end{figure*}

\begin{figure}[tp]
    \centering
    \includegraphics[width=\linewidth]{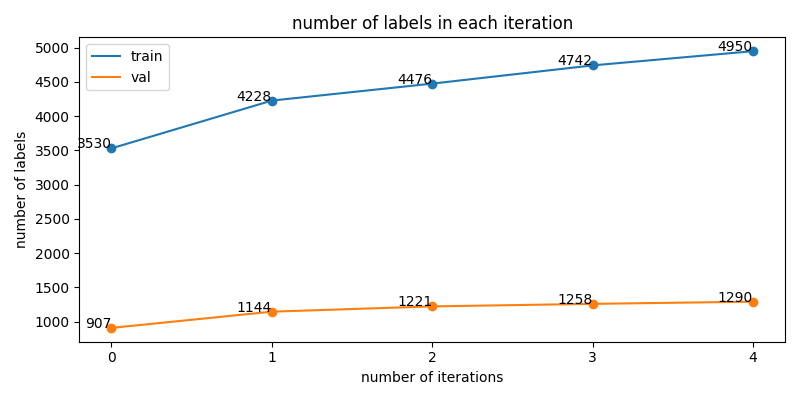}
    \caption{Total labels increase from 4437 to 6240 (40.6\% rise), with training and validation labels rising from 3530 to 4950 (40.2\% increase) and from 907 to 1290 (42.4\% increase), respectively.}
    \label{fig:fig3}
\end{figure}

\quad Here, we implement an iterative methodology to automate the data labeling process, followed by the subsequent application of NMS filtering techniques. This iterative methodology is pursued to acquire comprehensive, high-quality annotations tailored specifically for facial landmark detection tasks. This framework addresses the complexities of facial landmark detection and enhances diagnostic capabilities in medical imaging applications.
Figure~\ref{fig:fig2} illustrates the original landmark information and the labels generated through automated processes for each iteration across various datasets, namely 300W, AFW, HELEN, IBUG, and IFPW~\cite{sagonas2013300,sagonas2013semi,sagonas2016300,Zhu2012FaceDP,Le2012InteractiveFF}. These images exemplify the efficacy of automated data labeling across diverse racial demographics, thus showcasing the approach's adherence to demographic parity principles.
Simultaneously, while ensuring meticulous attention to label quality, we observe the number of labels for the original provided and each iterative stage, as illustrated in Figure~\ref{fig:fig3}. Notably, the number of labels increases from 4437 to 6240, reflecting a rise of approximately 40.6\%. To elaborate, the count of training and validation labels increases from 3530 to 4950 (a 40.2\% increase) and from 907 to 1290 (a 42.4\% increase), respectively.
\section{Experiments}
\label{sec:experiments}

\subsection{Experimental Setup}
\label{ssec:experimental_setup}

\quad Our proposed methodology will undergo evaluation through distinct avenues. We will assess its performance metrics, including precision, recall, average precision, and the MSE between the predicted and actual landmark coordinates provided in competition for all the landmarks detected in each image, by evaluating the best model attained at each developmental stage, as detailed in Section~\ref{ssec:comparison_with_baseline}. This approach ensures a thorough assessment of its performance capabilities. 
We aim to demonstrate our approach's practical applicability and effectiveness by disseminating our methods and engaging in this competitive environment. The results obtained from these evaluations will substantiate and validate our proposed methodology's efficacy in addressing challenges within facial landmark detection tasks.

\subsection{Implementation Details}
\label{ssec:implementation_details}

\quad In our implementation, we utilize the YOLOv8n-pose model exclusively for all comparative analyses. To ascertain the robustness of our validation procedures, we conducted all experiments using a laptop endowed with a 12th Gen Intel Core i7-12650H CPU, an NVIDIA GeForce RTX 4050 GPU, and 24GB of memory. This configuration ensures that we can comprehensively evaluate the performance of our proposed methodology. Furthermore, Table~\ref{tab:dataset} provides detailed descriptions of the datasets employed at different stages for the facial landmark detection task. This tabulated information elucidates the intricacies of our experimental setup and aids in understanding the nuances of our approach.
\begin{table}
\centering
  \caption{Count of images and labels for datasets at each iteration, with Iter0 indicating the label given at the beginning.}
  \label{tab:dataset}
  \begin{tabular}{lccccc}
    \toprule
    Datasets & Iter0 & Iter1 & Iter2 & Iter3 & Iter4 \\
    \midrule
    \# train & $3249$ & $3249$ & $3249$ & $3249$ & $3249$ \\
    \# val & $813$ & $813$ & $813$ & $813$ & $813$ \\
    \# total & $4062$ & $4062$ & $4062$ & $4062$ & $4062$ \\
    \cmidrule(lr){1-6}
    \# train labels & $3530$ & $4228$ & $4476$ & $4742$ & $5950$ \\
    \# val labels & $907$ & $1144$ & $1221$ & $1258$ & $1290$ \\
    \# total labels & $4437$ & $5372$ & $5697$ & $6000$ & $6240$ \\
    \bottomrule
  \end{tabular}
\end{table}

\subsection{Comparison with baseline}
\label{ssec:comparison_with_baseline}

\quad In this section, we utilize metrics from the Iter0 dataset as our baseline for comparison, enabling an analysis of performance metrics across successive iterations of enhanced labeling datasets (Iter1 through Iter4), outlined in Table~\ref{tab:baseline}. Implementing the proposed technique results in a notable improvement in the MSE score, a vital indicator of variance between predicted and actual values.
\begin{table}
\centering
  \caption{Comparison of precision, recall, average precision, and MSE for datasets at each iteration.}
  \label{tab:baseline}
  \begin{tabular}{lccccc}
    \toprule
    Datasets & Iter0 & Iter1 & Iter2 & Iter3 & Iter4 \\
    \midrule
    \textbf{Precision (\%)} & $91.5$ & $93.2$ & $95.3$ & $94.1$ & $94.1$ \\
    \textbf{Recall (\%)} & $92.7$ & $92.8$ & $91.8$ & $90.5$ & $90.0$ \\
    \textbf{AP}$_{50}^{val}$ \textbf{(\%)} & $95.9$ & $94.2$ & $93.4$ & $91.7$ & $91.3$ \\
    \textbf{AP}$_{50:95}^{val}$ \textbf{(\%)} & $68.4$ & $67.4$ & $66.1$ & $63.9$ & $62.8$ \\
    \textbf{MSE} & $17.72$ & $18.21$ & $18.43$ & $18.79$ & $\boldsymbol{18.81}$\\
    \bottomrule
  \end{tabular}
\end{table}



\section{Conclusion}
\label{sec:conclusion}

\quad In this paper, we propose an iterative refinement strategy designed for automated data labeling, presenting compelling evidence of its remarkable effectiveness in revolutionizing performance within the realm of landmark detection tasks. Beyond its stellar performance in facial landmark detection, our approach seamlessly adapts to various tasks, including finger joint localization, nuanced body pose estimation, and intricate polygon segmentation. This remarkable adaptability underscores the versatility and scalability of our approach across myriad domains, necessitating precise landmark localization. Moreover, our methodology facilitates swift and accurate diagnosis of large populations, offering a vital lifeline in resource-constrained environments by enabling early symptom detection and fostering societal well-being.


{\small
\bibliographystyle{ieeenat_fullname}
\bibliography{11_references}
}

\ifarxiv \clearpage \appendix \section{Appendix Section}
Supplementary material goes here.
 \fi

\end{document}


\title{\paperTitle}
\author{\authorBlock}
\maketitlesupplementary

\section{Appendix Section}
Supplementary material goes here.

{\small
\bibliographystyle{ieee_fullname}
\bibliography{11_references}
}